# APLICAÇÃO DE ABP NO DESENVOLVIMENTO E MODELAGEM DAS CINEMÁTICAS DE MANIPULADORES ROBÓTICOS COM INTERDISCIPLINARIDADE ENTRE PROJETO ASSISTIDO POR COMPUTADOR, ROBÓTICA E MICROCONTROLADORES

**Afonso Henriques Fontes Neto Segundo**
Mestre em Ciência da Computação pela Universidade Estadual do Ceará
Professor na Universidade de Fortaleza (UNIFOR)
Endereço: Av. Washington Soares, 1321 – Edison Queiroz, Fortaleza – CE, Brasil
E-mail: afonsof@unifor.br

**Joel Sotero da Cunha Neto**
Mestre em Informática Aplicada pela Universidade de Fortaleza
Professor na Universidade de Fortaleza (UNIFOR)
Endereço: Av. Washington Soares, 1321 – Edison Queiroz, Fortaleza – CE, Brasil
E-mail: joelsotero@unifor.br

**Paulo Cirillo Souza Barbosa**
Engenheiro de Controle e Automação pela Universidade de Fortaleza (UNIFOR)
Pesquisador na Universidade de Fortaleza (UNIFOR)
Endereço: Av. Washington Soares, 1321 – Edison Queiroz, Fortaleza – CE, Brasil
E-mail: pauloc@unifor.br

**Raul Fontenele Santana**
Graduando em Engenheiro de Controle e Automação pela Universidade de Fortaleza (UNIFOR)
Pesquisador na Universidade de Fortaleza (UNIFOR)
Endereço: Av. Washington Soares, 1321 – Edison Queiroz, Fortaleza – CE, Brasil
E-mail: raulfontenele@edu.unifor.br

***Resumo:*** *Tendo em vista a dificuldade de alunos em calcular as cinemáticas direta e inversa de um manipulador robótico utilizando apenas ferramentas convencionais de uma sala de aula, este artigo propõe a aplicação de Aprendizagem Baseada em Projeto (ABP) através da concepção, desenvolvimento, modelagem matemática de um manipulador robótico como projeto integrador das disciplinas de Robótica Industrial, Microcontroladores e Projeto Assistido por Computador com alunos da Engenharia de Controle e Automação da Universidade de Fortaleza. Uma vez projetado e usinado, o braço manipulador foi montado utilizando servomotores conectados a uma placa de prototipagem microcontrolada, para ter então sua cinemática calculada. Ao fim são apresentados os resultados que o projeto trouxe para o aprendizado das disciplinas sobre a ótica do professor orientador e dos alunos.*

***Palavras-Chave:*** *ABP. Interdisciplinaridade. Projeto Assistido por Computador. Microcontroladores. Robótica Industrial. Cinemáticas direta e inversa. Manipulador Robótico.*

# 1 INTRODUÇÃO

A preparação de novos engenheiros para o mercado de trabalho é um árduo trabalho para educadores, tendo em vista que vivemos em um período onde em poucos anos, grandes mudanças podem acontecer como por exemplo na economia mundial ou na tecnologia dominante. Logo, os educadores têm se questionado quais mudanças devem ocorrer no contexto escolar, principalmente na prática docente, fazendo com que os recém graduados possam atender a demanda do mercado de trabalho (BARBOSA; MOURA,2014).

Para Moura(2014), os estudantes de engenharia devem ensinados através de uma aprendizagem contextualizada e orientada para o uso das tecnologias. Além disso, é de grande importância que os futuros engenheiros também tenham desenvolvido habilidades em resolver problemas, conduzir projetos e que sejam capazes de desempenhar condições de formação humana como por exemplo, iniciativa, liderança, proatividade, entre outros. Logo, a utilização de metodologias ativas no processo de aprendizado, contribuem na imersão dos alunos em um ambiente de aprendizagem contextualizado e de grande relevância para formação de um Engenheiro. Neste ambiente, o professor pode atuar como orientador, supervisor ou até mesmo como facilitador do processo de aprendizagem, tendo em vista que o ponto importante da metodologia ativa é fazer com que os alunos exercitem suas habilidades de raciocinar, observar, refletir, criar competitividade saudável, criar soluções diferentes de um mesmo problema (BARBOSA,2003) .

O início do projeto se deu por uma pesquisa realizada em sala de aula com os alunos, para que fosse possível avaliar alguns tópicos como, semestre do aluno, auto avaliação na disciplina e se o projeto do manipulador físico é um fator auxiliador no ensino. Com esses dados podemos notar que dos 13 alunos, 84.6% aprovam a metodologia ativa como forma de ensino como ilustra a Figura 1. Além disto com a Figura 1 podemos notar que 100% dos alunos estão perto da conclusão do curso.

Figura 1: Primeira pesquisa.

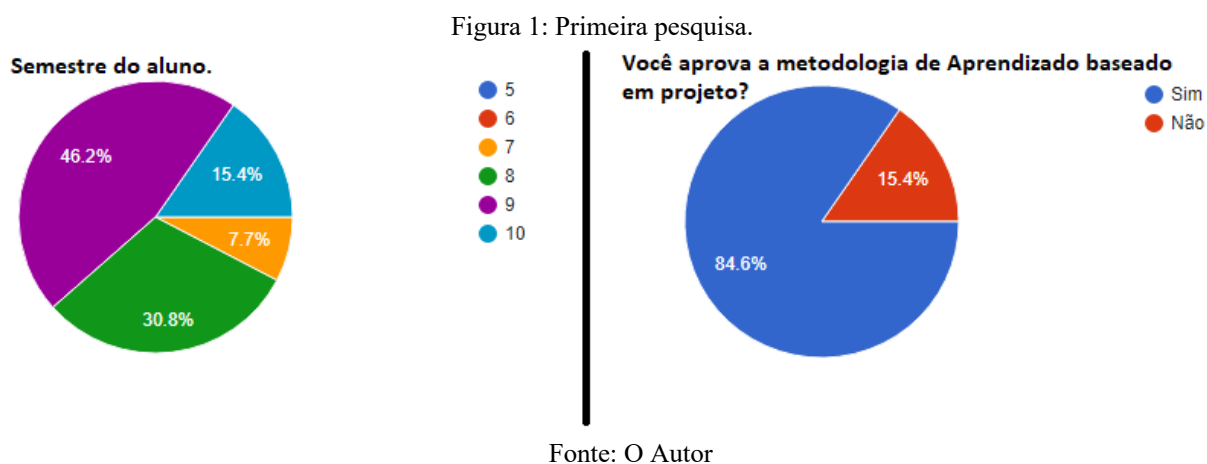

Fonte: O Autor

Com a pesquisa, foi possível expressar em gráficos também, qual o desempenho atual do aluno na disciplina e se por acaso o projeto pudesse ampliar sua capacidade de aprendizado como ilustra a Figura 2.

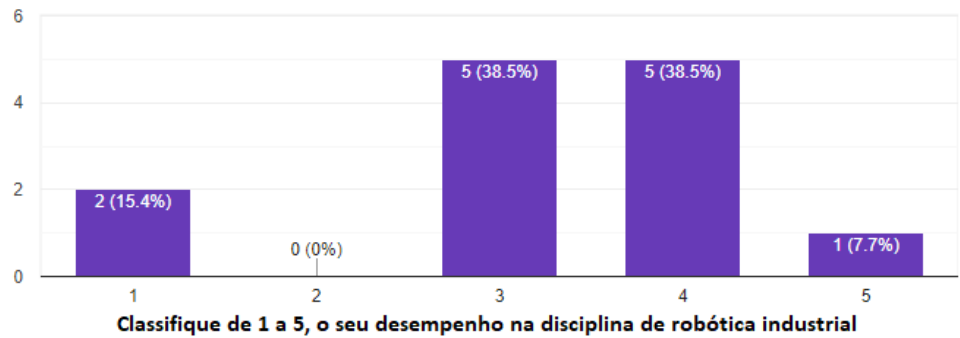

Figura 2: Segunda pesquisa.

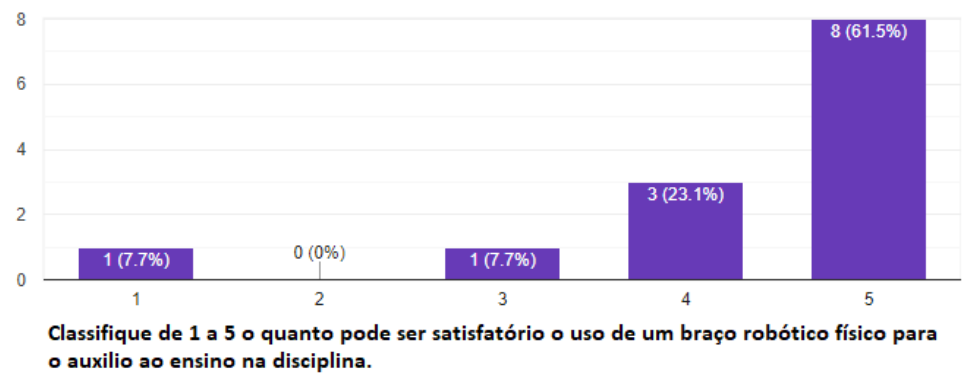

Fonte: O Autor

Baseado neste contexto, foi proposto pelo professor da disciplina de robótica industrial da Universidade de Fortaleza, o desenvolvimento de um manipulador robótico que faria integração com outras disciplinas vistas durante o curso como Projeto Assistido por Computadores, Algoritmos e Programação de Computadores. O presente artigo, se limita ao desenvolvimento do manipulador via Computer Aided Design(CAD), sua montagem, implementação de modelos matemáticos das cinemáticas direta e inversa e desenvolvimento de algoritmos com intuito de proteção, para evitar possíveis colisões com o próprio manipulador. No futuro, este projeto se estenderá a união de mais três manipuladores produzidos por outros alunos para compor um robô quadrúpede.

**DESENVOLVIMENTO DO MANIPULADOR ROBÓTICO.**

Para o desenvolvimento do manipulador robótico, o professor da disciplina estipulou alguns objetivos específicos que são indispensáveis para compor a nota dos alunos:

1. Desenvolver um manipulador robótico com três graus de liberdade.
2. Realizar a fabricação e montagem do mesmo por um processo de baixo custo.
3. Modelagem de equações matemáticas das cinemáticas direta e inversa.
4. Implementação de algoritmos em microcontrolador, o qual atuará no movimento do manipulador pelas equações do item anterior.
5. Implementar algoritmo de proteção para evitar colisões do manipulador com ele mesmo e com a superfície de atuação.

## 1.1 Projeto de manipulador via CAD e fabricação

Partindo dos objetivos específicos e com o conhecimento adquirido na disciplina de Projeto Assistido por Computador, que agrega ao aluno o conhecimento de modelar peças tridimensionalmente em software a partir de um modelo real ou de uma ideia de projeto. Foi então, desenvolvido no software Solid Works cada uma das peças que compõem o manipulador levando em consideração os 3 graus de liberdade como ilustra a Figura 3. Em seguida, foi concordado entre os alunos que o processo de fabricação que viabiliza o projeto, seria através de corte a laser, que utiliza as medidas das peças modeladas para cortar madeira do tipo, com o mesmo formato e escala.

Figura 3: Projeto modelado com três graus de liberdade.

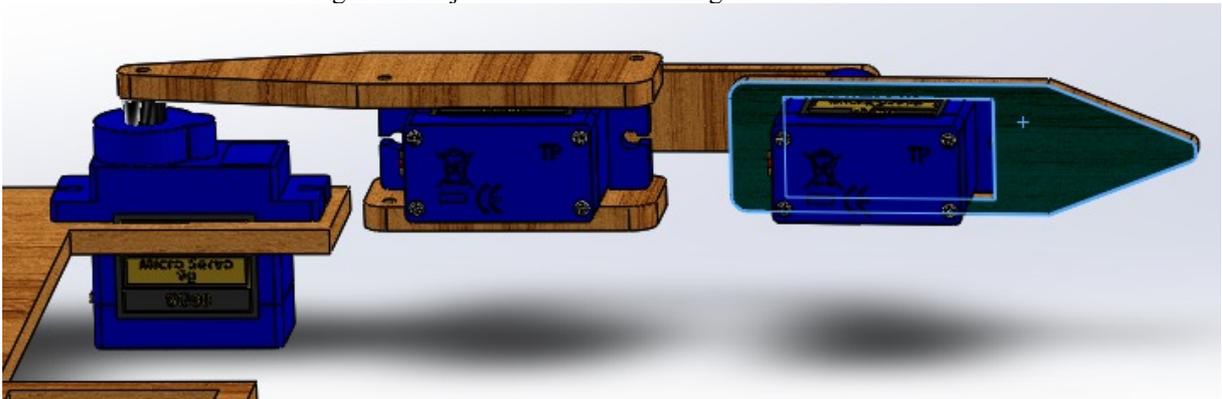

Fonte: O Autor

## 1.2 Modelagem de cinemáticas.

Para (CRAIG,2012), os robôs utilizados em ambientes industriais, devem executar suas tarefas de acordo com sua ferramenta localizada em sua extremidade. Deste ponto, pode-se criar coordenadas em espaço cartesiano definindo sua posição no espaço, para que seja possível a implementação de algoritmos matemáticos que possam converter esses valores nos deslocamentos de cada grau de liberdade do manipulador robótico, este processo é chamado de cinemática inversa.

Com esse conjunto de angulações, deve-se aplicar um algoritmo de cinemática direta que auxilia o controlador ter noção onde cada junta está posicionada. Logo, esse processo, têm como parâmetros os valores de ângulos e é calculado a posição e orientação do elemento da extremidade do robô(ROSÁRIO,2005).

### 1.1.1. Aplicação de Cinemática Direta

Inicialmente, para se aplicar a cinemática direta, é preciso posicionar frames em cada articulação do manipulador ilustrados na Figura 4 como coordenadas 0 até 4. Com os frames posicionados, deve-se expressar através de modelagem matemática, todas as possíveis rotações e translações para que se encontre o valor das coordenadas X,Y e Z onde a extremidade do manipulador se encontra, ou seja, partindo do frame zero inicial, deve-se expressar através de matriz o necessário para chegar ao frame 4. Este processo pode ser representado pela matriz

M04 a qual é encontrada pela Equação 1.0, onde cada elemento dessa equação pode ser encontrado pelas Equações 1.1 a 1.4.

Figura 4: Frames posicionados.

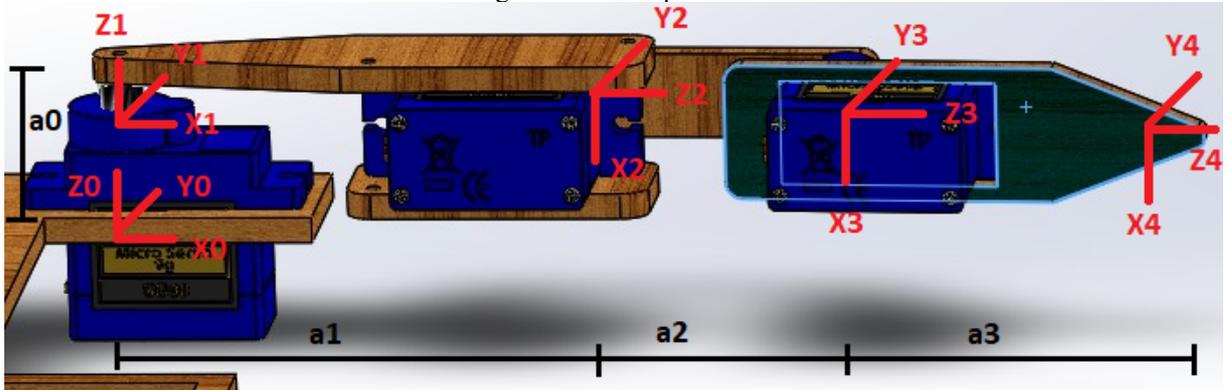

Fonte: O Autor

$$M04 = M01 * M12 * M23 * M34 \qquad (1.0)$$

$$M01 = Trans('Z', a0) \qquad (1.1)$$

$$M12 = Rot('Z', \theta 1) * Rot('Y', pi/2) * Trans('Z', a1) * Rot('Y', \theta 2) \qquad (1.2)$$

$$M23 = Trans('Z', a2) * Rot('Y', \theta 3) \qquad (1.3)$$

$$M34 = Trans('Z', a3) * Rot('Y') \qquad (1.4)$$

Com os cálculos realizados, e a matriz M04 já calculada, é possível extrair as coordenadas onde a extremidade do manipulador se encontra, pois os elementos da quarta coluna e nas três primeiras linhas, são respectivamente X, Y e Z.

### 1.1.2. Aplicação de Cinemática Inversa

Para cinemática inversa, é necessário modelar as equações matemáticas a partir dos valores das coordenadas X,Y e Z da ponta da ferramenta, a fim de encontrar os valores dos ângulos de rotação ou sua translação necessária para que fosse possível chegar até X,Y e Z. Partindo de uma vista superior da ferramenta ilustrada pela Figura 5, é possível utilizar o teorema de pitágoras como mostra a Equação 1.4 para encontrar o valor de w. Em seguida é possível utilizar a relação trigonométrica para encontrar o valor do primeiro ângulo de articulação do manipulador representado pela Equação 1.5.

Figura 5: Vista Superior.

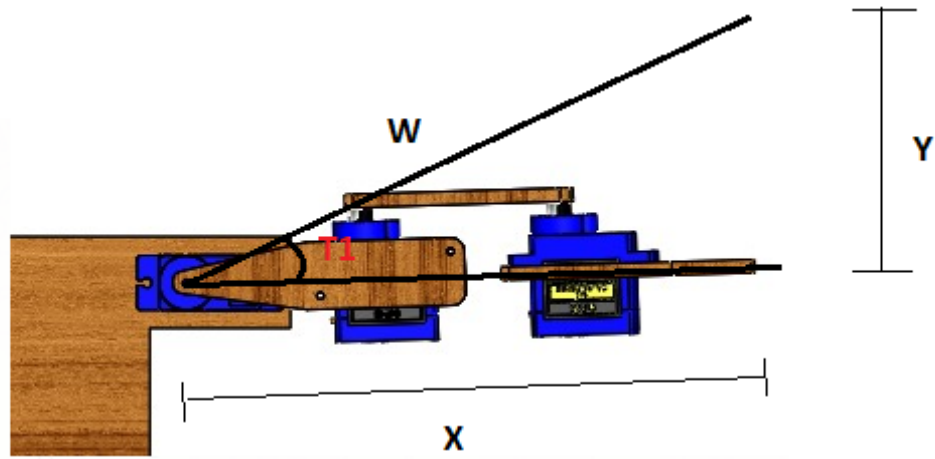

Fonte: O Autor

$$W = \sqrt{X^2 + Y^2} \qquad (1.4)$$

$$T1 = atan\left(\frac{Y}{X}\right) \qquad (1.5)$$

Como não é possível encontrar os outros valores de ângulos de articulações pela vista superior, é necessário a utilização da vista lateral para modelar as equações como ilustra a Figura 7. Consideramos aqui, que caso $T1$ tenha seu valor alterado, a vista acompanhará o manipulador, fazendo com que o valor de w, sempre seja visualizado e não uma projeção do mesmo. Com isto, podemos traçar uma reta do eixo de $θ1$ até a extremidade da ferramenta formando um triângulo retângulo com a4-Z e w. Deste ponto, como é verificado na Figura 6, é possível extrair as equações 1.6 a 1.x, para se encontrar os valores de $T2$ e $T3$.

Figura 6: Vista Lateral.

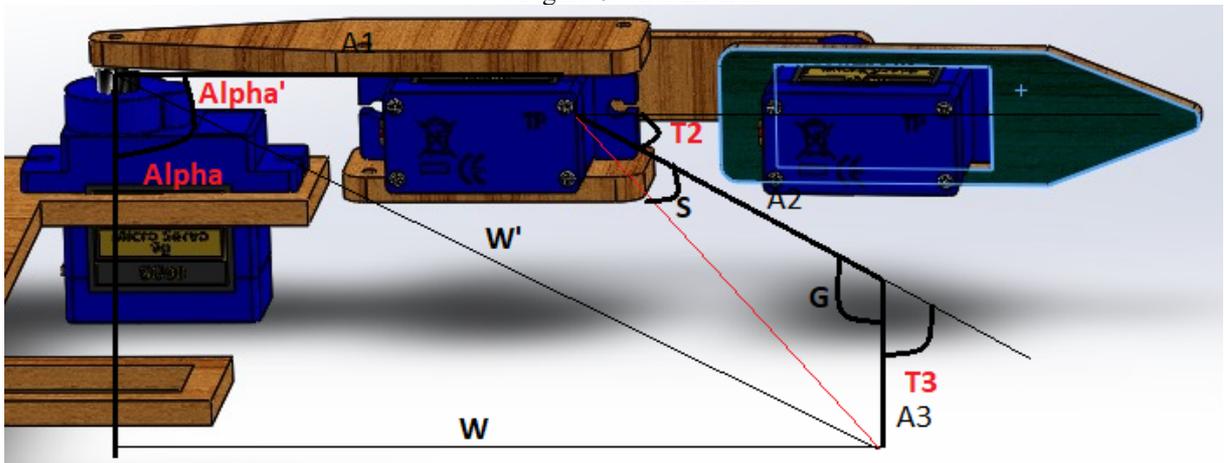

Fonte: O Autor

$$w' = \sqrt{w^2 + (a0 - Z)^2} \qquad (1.6)$$

$$\alpha = atan\left(\frac{w}{(a0-Z)}\right) \qquad (1.7)$$

$$\alpha' = (\pi/2) - \alpha \qquad (1.8)$$

$$b == \sqrt{a1^2 + w'^2 - 2*a1*w'*cos(\alpha')} \qquad (1.9)$$

$$\alpha'' = asin(w' * sin(\frac{\alpha'}{b})) \tag{2.0}$$

$$\gamma = acos(\frac{(a2^2 + a3^2 - b^2)}{2*a2*a3}) \tag{2.1}$$

$$\gamma' = asin(\frac{(a3*sin(\gamma))}{b}) \tag{2.2}$$

$$T2 = \pi - \alpha'' - \gamma' \tag{2.3}$$

$$T3 = \pi - \gamma \tag{2.4}$$

## 2 Validação dos cálculos

Para validar se as duas modelagens foram implementadas corretamente, é necessário partir das coordenadas x,y,z de onde a extremidade do manipulador se encontra e aplicar esses valores às equações de cinemática inversa. Para que seja possível encontrar valores dos ângulos das articulações que ocasionaram a extremidade do manipulador chegar nas coordenadas. Em seguida, aplicamos a modelagem de matrizes da cinemática direta com os valores dos ângulos obtidos na inversa da base até a extremidade da ferramenta. A comprovação dos valores é visualizada na Figura 7, que ilustra os valores das coordenadas (3,-5,-8) aplicados na cinemática inversa, e mostra que aplicando os valores de ângulos na cinemática direta, é possível visualizar o valor de coordenadas com poucas casas decimais de diferença.

Figura 7: Validação de dados.

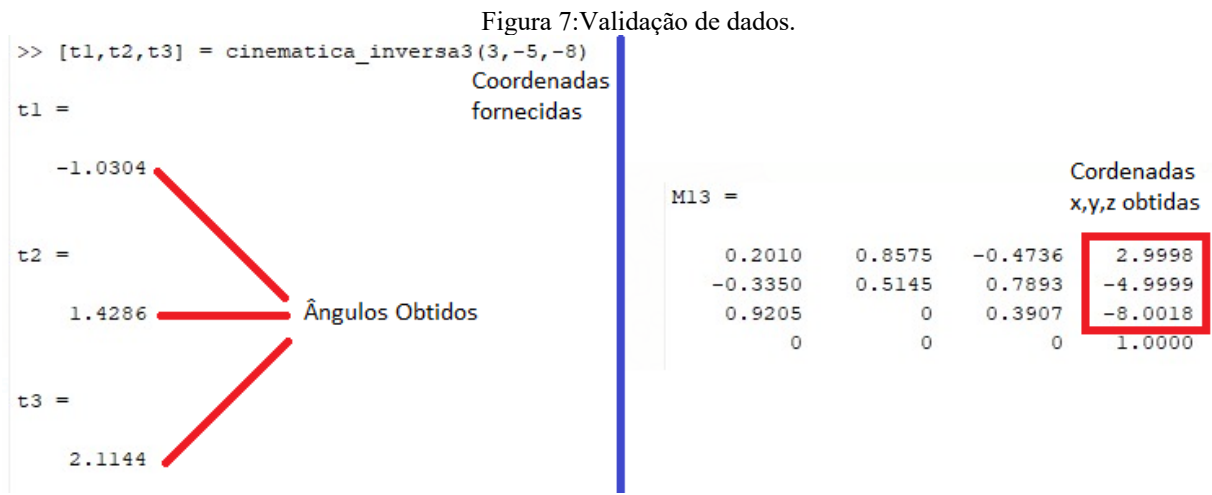

Fonte: O Autor

## 3 Implementação de algoritmos.

Com as equações matemáticas modeladas, o seguinte passo é utilizá-las para desenvolver um algoritmo, o qual aplicado em um microcontrolador, pode realizar diferentes atuações nos servomotores, dando movimento ao manipulador robótico. Logo, esta fase do projeto engloba outras duas disciplinas vistas na graduação do curso de Engenharia de Controle e Automação. Na disciplina de Microcontroladores, onde é ensinado os princípios de utilização de alguns microcontroladores, sua vista de hardware e como desenvolver algoritmos para manipular suas entradas e saídas.

Considerando que o manipulador não pode realizar todo tipo de movimento, já que sua extremidade ou outra parte do manipulador não pode colidir com o mesmo, e não poderá também colidir com a base onde está fixado. Foi utilizado uma técnica aprendida na disciplina de Algoritmos e Programação de Computadores, que se baseia em limitar a movimentação do braço em certas coordenadas x,y e z para que não ocorra nenhuma colisão. Então, foi definido entre os alunos da equipe que o controlador irá esperar como dado de entrada a coordenada desejada para que a extremidade do manipulador se desloque. A Figura 8 ilustra com um retângulo vermelho a região onde o manipulador não atuará.

Figura 8:Limite de atuação.

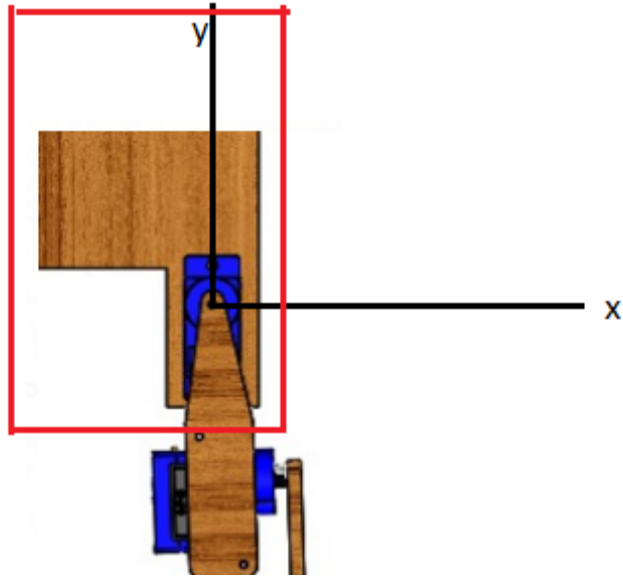

Fonte: O Autor

Foi definido e a distância de atuação em Z partindo do servo até o solo no valor de 60mm. Logo, a primeira proteção é limitar se o usuário informar algum valor menor que -60, o programa limite para -60. Para as coordenadas X e Y foi definido duas possibilidades, se por acaso o valor de Y fosse menor que zero e X menor que -51 o valor de X é definido para -51. Caso Y seja positivo e X maior que 53 o valor em X é redefinido para 52. Em fase de teste, foi verificado que para certas coordenadas os valores dos ângulos obtidos pela cinemática inversa estavam errados, isso se deu pelo fato da divisão das equações 2.1 e 2.2 retornarem valores menores que 1 ou maiores que 1. Foi pesquisado que se este problema acontecer, basta retirar a parte inteira do número e seguir com os cálculos somente com os a parte flutuante do mesmo. O código fonte do controlador e os projetos de cada parte do manipulador podem ser encontrados no GitHub https://github.com/raulfontenele/Robotic-Arm

## 4 Resultados e discussões.

Com o manipulador finalizado, foi realizado uma nova pesquisa em sala de aula para avaliar o nível de aprendizado obtido pelos alunos. Baseado na pesquisa foi possível observar pelas Figura 10 e 11 que a utilização de metodologia ativa pelo projeto foi satisfatória, tanto no desempenho dos alunos que aumentou consideravelmente, como pelo ambiente onde os alunos estão imersos, onde foi possível compartilhar conhecimentos entre integrantes da equipe com conhecimentos adquiridos na graduação. Além do conhecimento adquirido com o projeto, foi possível notar que algumas habilidades interpessoais como, espírito de equipe, proatividade, liderança, capacidade de resolver problemas foram bem desenvolvidas no decorrer do projeto. Por fim, a utilização de metodologia ativa no ensino foi considerada satisfatória pela maioria dos alunos e pôde ser um fator de integração de aprendizado entre alunos e professor.

Figura 10:Algoritmo Implementado.

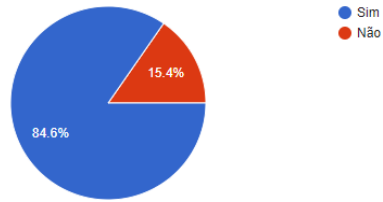

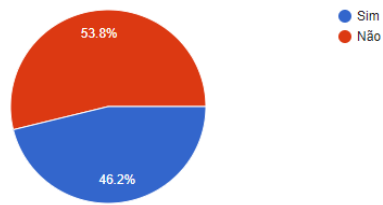

Fonte: O Autor

Figura 10:Algoritmo Implementado.

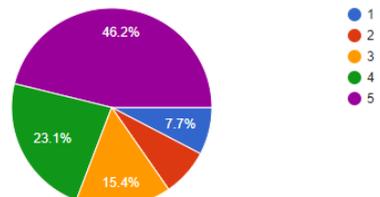

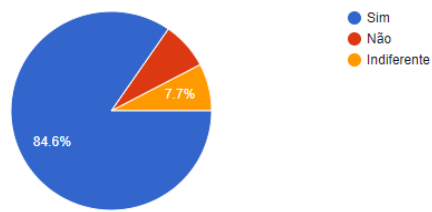

Fonte: O Autor

## 5   Trabalhos Futuros

Com o manipulador robótico finalizado, o trabalho futuro é a união de outros manipuladores idênticos ao desenvolvido, em uma base dispostos de tal de tal maneira que formem um quadrúpede como ilustra a Figura 11. O desafio nesta fase do projeto, é unir as equipes da sala para que desenvolvam uma nova cinemática para que o conjunto de manipuladores possa se deslocar para um sentido desejado, levando em consideração que nenhum dos manipuladores possam se chocar uns com os outros.

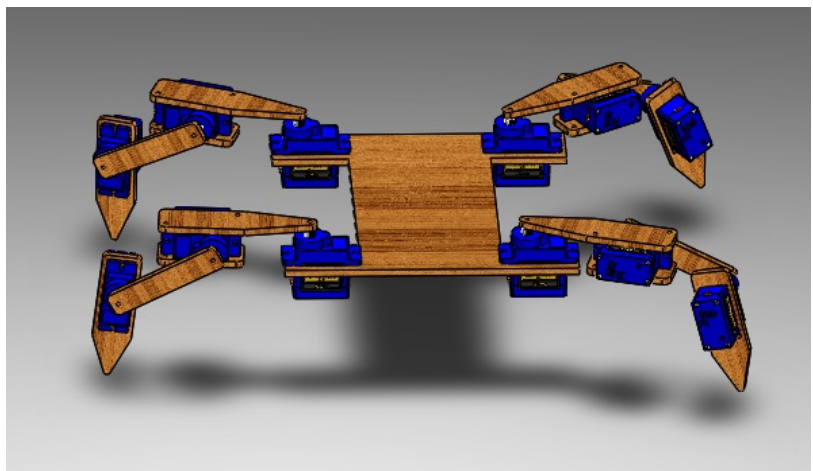

Fonte: O Autor

## REFERÊNCIAS

# APPLICATION OF ABP IN THE DEVELOPMENT AND MODELING OF KINEMATICS OF ROBOTIC MANIPULATORS WITH INTERDISCIPLINARITY BETWEEN COMPUTER, ROBOTIC AND MICROCONTROLLER ASSISTED PROJECT

*Abstract: Considering the difficulty of students in calculating the direct and inverse kinematics of a robotic manipulator using only conventional tools of a classroom, this article proposes the application of Project Based Learning (ABP) through the design, development, mathematical modeling of a robotic manipulator as an integrative project of the disciplines of Industrial Robotics, Microcontrollers and Computer Assisted Design with students of the Control and Automation Engineering of the University of Fortaleza. Once designed and machined, the manipulator arm was assembled using servo motors connected to a microcontroled prototyping board, to then have its kinematics calculated. At the end are presented the results that the project has brought to the learning of the disciplines on the optics of the tutor and students.*

**Key-words:** *ABP. Interdisciplinarity. Computer Assisted Design. Microcontrollers. Industrial Robotics. Direct and inverse kinematics. Robotic manipulator.*